\documentclass[conference]{IEEEtran}
\IEEEoverridecommandlockouts
\usepackage{cite}
\usepackage{amsmath,amssymb,amsfonts}
\usepackage{algorithmic}
\usepackage{graphicx}
\usepackage{textcomp}
\usepackage{xcolor}
\usepackage{multicol}
\usepackage{algorithm}
\usepackage{algorithmic}
\usepackage{multirow} 
\usepackage{array}
\usepackage{float}
\usepackage[table,xcdraw]{xcolor}
\usepackage[hidelinks]{hyperref} 
\def\BibTeX{{\rm B\kern-.05em{\sc i\kern-.025em b}\kern-.08em
    T\kern-.1667em\lower.7ex\hbox{E}\kern-.125emX}}
\begin{document}

\title{A Bidirectional Gated Recurrent Unit Model for PUE Prediction in Data Centers%
\thanks{© 2025 IEEE. This is the author’s accepted version of the work published in the Proceedings of the 2025 International Joint Conference on Neural Networks (IJCNN).
D. D. Kannan, A. Trivedi, and D. Srinivasan, “A Bidirectional Gated Recurrent Unit Model for PUE Prediction in Data Centers,” IJCNN 2025, Rome, Italy, pp. 1–8, 2025.
DOI: 10.1109/IJCNN64981.2025.11227238.
The final published version is available at: https://ieeexplore.ieee.org/document/11227238}}

\author{\IEEEauthorblockN{1\textsuperscript{st} Dhivya Dharshini Kannan}
\IEEEauthorblockA{\textit{Department of Electrical and Computer} \\
\textit{Engineering} \\
\textit{National University of Singapore}\\
Singapore \\
e1322630@u.nus.edu}
\and
\IEEEauthorblockN{2\textsuperscript{nd} Anupam Trivedi}
\IEEEauthorblockA{\textit{Department of Electrical and Computer } \\
\textit{Engineering} \\
\textit{National University of Singapore}\\
Singapore \\
eleatr@nus.edu.sg}
\and
\IEEEauthorblockN{3\textsuperscript{rd} Dipti Srinivasan}
\IEEEauthorblockA{\textit{Department of Electrical and Computer } \\
\textit{Engineering} \\
\textit{National University of Singapore}\\
Singapore \\
dipti@nus.edu.sg}
}

\maketitle

\begin{abstract}
Data centers account for significant global energy consumption and a carbon footprint. The recent increasing demand for edge computing and AI advancements drives the growth of data center storage capacity. Energy efficiency is a cost-effective way to combat climate change, cut energy costs, improve business competitiveness, and promote IT and environmental sustainability. Thus, optimizing data center energy management is the most important factor in the sustainability of the world. Power Usage Effectiveness (PUE) is used to represent the operational efficiency of the data center. Predicting PUE using Neural Networks provides an understanding of the effect of each feature on energy consumption, thus enabling targeted modifications of those key features to improve energy efficiency. In this paper, we have developed Bidirectional Gated Recurrent Unit (BiGRU) based PUE prediction model and compared the model performance with GRU. The data set comprises 52,560 samples with 117 features using EnergyPlus, simulating a DC in Singapore. Sets of the most relevant features are selected using the Recursive Feature Elimination with Cross-Validation (RFECV) algorithm for different parameter settings. These feature sets are used to find the optimal hyperparameter configuration and train the BiGRU model. The performance of the optimized BiGRU-based PUE prediction model is then compared with that of GRU using mean squared error (MSE), mean absolute error (MAE), and R-squared metrics.

\end{abstract}


\section{Introduction}

The capacity of data centers (DC) in Southeast Asia is expected to increase by 19\% annually between 2021 and 2026 due to the present focus on edge computing \cite{b1}. By 2030, the global DC market will grow to reach US\$554.4 billion, with Indonesia, Malaysia, and Singapore serving as the main development areas \cite{b2}. Singapore intends to increase its current 1.4 GW capacity of more than 70 DCs by 300 MW. The goal established by the Singaporean government is for all DCs in the country to have a PUE of 1.3 or below at 100\% IT load \cite{b3}. 

By 2027, the demand for DC storage capacity is expected to double as a result of advances in AI \cite{b4}. According to the International Energy Agency, in 2022, DCs consumed about 460 TWh of electricity, which is 2\% of the global electricity consumption \cite{b5} and equivalent to the total electrical usage of Japan \cite{b6}. By 2026, DC energy consumption is expected to be about 1000 TWh. By 2027, 75\% of the companies will have established sustainable data center initiatives, up from only 5\% in 2022, according to Gartner \cite{b7}.

Power Usage Effectiveness (PUE) is the direct representation of energy consumed by DCs and is the ratio of total DC energy usage to IT equipment energy usage \cite{b8}. Modern DCs are complex systems with huge amounts of interdependent data from various sensors such as IT load, cooling system, environmental parameters, etc, which makes it difficult to estimate PUE using traditional mathematical methods.  Predefined feature interactions are not necessary for neural networks (NN). It creates the best-fit model by automatically looking for patterns and interactions between features \cite{b9}. Thus, capturing the complex inter-dependencies between features and their impact on PUE. 

Various PUE prediction models based on Neural Networks \cite{b9}\cite{b10}, Deep Neural Network \cite{b11}\cite{b12}, Gated Recurrent Unit \cite{b13}, Multilayer Perceptron, Resilient Backpropagation-based Deep Neural Network, and Attention-based Long-Short Term Memory \cite{b14} were proposed in previous studies.

RNN (Recurrent Neural Network) is a type of NN applied for time-series-based data for sequence prediction. It maintains a hidden state that is updated at each time step to keep the information from the previous steps in memory \cite{b15}. Since the data set for the PUE prediction is time-series based, RNN is the best choice for developing the PUE prediction model. 

Gated Recurrent Unit (GRU) is a variant of RNN that can handle time-dependent sequential data by considering both past and present input. It overcomes the gradient vanishing problem of RNN by using two gates, namely the forget gate and update gate by controlling the amount of past data to discard and retain, respectively \cite{b13}. GRU uses a recurrent structure to store and retrieve information and does not account for future state. This may lead to limited prediction accuracy. 

The Bidirectional Gated Recurrent Unit (BiGRU) overcomes this limitation by utilizing a future layer which allows predicting data sequence in the opposite direction \cite{b16}. BiGRU is an upgraded from of GRU that uses two hidden states \cite{b17}. Thus, BiGRU-based models have the potential to outperform GRU-based models because of their ability to handle past and future data.

\begin{figure*}[ht]
    \centering
    \includegraphics[width=0.73\linewidth]{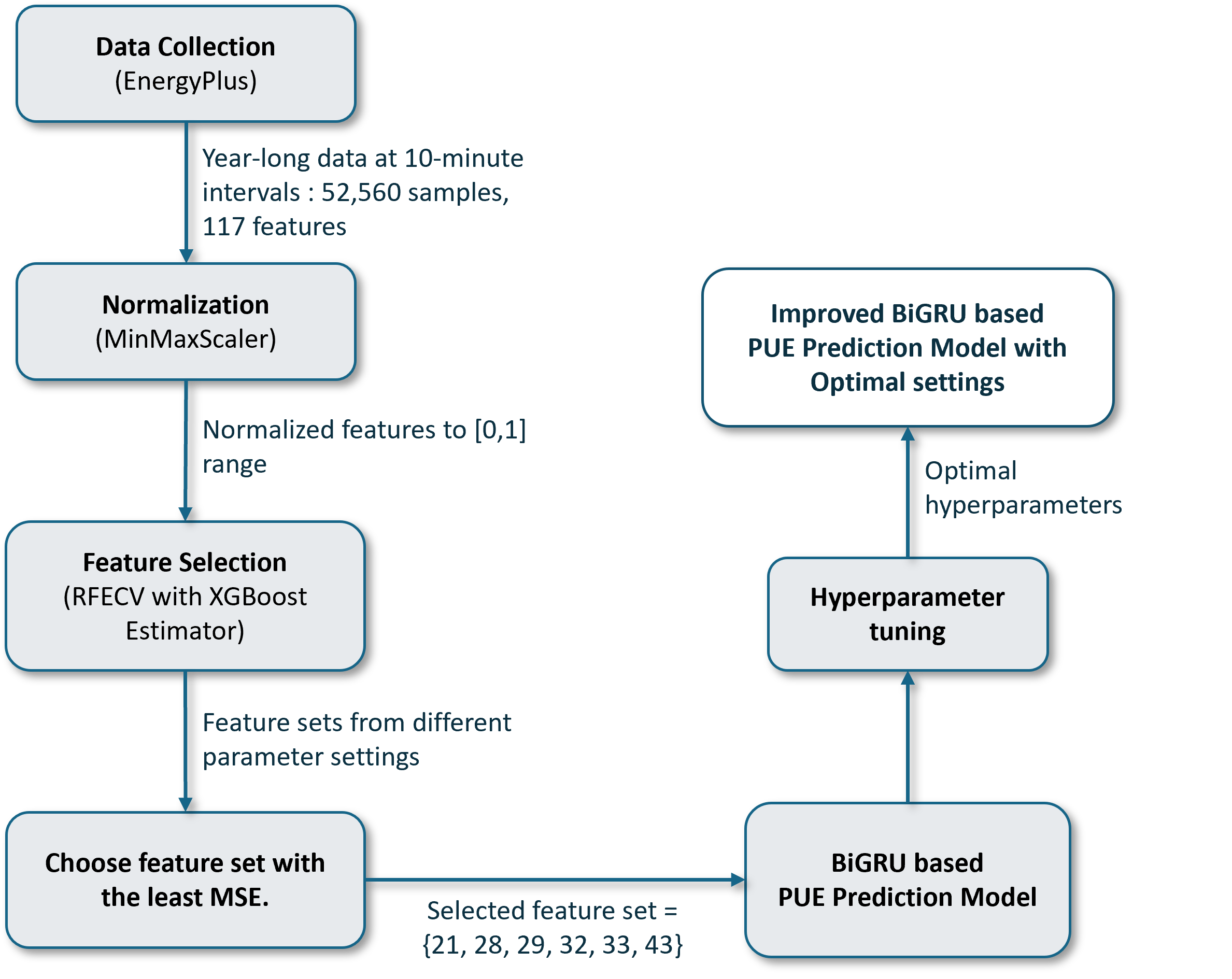}
    \caption{Flowchart of steps involved in the development of PUE prediction model}
    \label{fig1}
\end{figure*}

The main contributions of our research are as follows:
\begin{itemize}
    \item A high performing BiGRU-based PUE prediction model is developed using just 21 most relevant features. 
    \item The dataset is generated for over a year at 10-minute intervals using the EnergyPlus platform, simulating a DC in Singapore by utilizing Singapore weather data to obtain dataset of 52,560 samples with 117 features.
    \item To identify the most influential features, we used the RFECV algorithm using XGBoost as an estimator, and MSE as the scoring parameter. This reduced the feature set to key attributes, with the number of selected features being \{21, 28, 29, 32, 33, 43\}, corresponding to the minimum MSE.
    \item Hyperparameter tuning for each feature set was carried out to construct an improved BiGRU-based PUE prediction model with optimal configurations. The proposed BiGRU model can predict PUE more accurately than the GRU model, achieving a very low MAE of 0.00038. 
\end{itemize}

Section II provides information on related works. Section III discusses the details of our proposed approach for the PUE prediction model development, such as data set generation, data pre-processing, feature selection, and BiGRU based model. Section IV includes the experiment and performance evaluation for various parameter settings of BiGRU. Finally, the conclusion is presented.  

\section{Related works}

Green Grid first developed the PUE as a standard metric to relate the total energy consumed to the energy used by the servers for measuring DC energy efficiency \cite{b8}. PUE is a valuable tool to optimize the energy consumption and minimize carbon footprint of DCs, with PUE close to 1 is considered good. Due to the growing demand for DCs, a lot of research has been done on how to maximize DC energy efficiency by developing a PUE prediction model using machine learning techniques.

In order to estimate PUE in DCs, Gao et al. \cite{b9} suggested a NN framework obtaining an MAE of 0.004 ± 0.005 for a PUE of 1.1. Their model was validated and applied at Google DCs to simulate process water supply temperature changes, catch error in machine readings, and to optimize DC parameters with the help of local expertise. This work showcased the potential of NN in utilizing the sensor data to enhance energy efficiency in complex DC systems. 

Yang et al. \cite{b10} applied NN with gradient boosting to optimize cooling system settings, at Tencent Tianjin which is one of China's largest hyperscale DCs. They compared the performance of PUE prediction models using different NN models, with their proposed NN achieving an MAE of 0.0259\% using 38 features. They used this model to perform Sensitivity Analysis which led to PUE reduction of 0.005 with an annual energy saving of about 1500 MWh.

Ounifi et al. \cite{b11} used a DNN-based model for the prediction of PUE and used the root mean square error (RMSE) as the evaluation metric achieving RMSE of 0.00174. In \cite{b14}, Ounifi et al. explored other machine learning models, including MLP, Resilient Backpropagation-based DNN, and Attention-based LSTM. They tested these models on datasets from 2 real DCs with Attention-based LSTM achieving the least MAE of 0.0026 and 0.0039 respectively.

Hu et al. \cite{b12} focused on refrigeration systems as it accounts for over 40\% of total DC energy consumption. They employed a DNN to develop a PUE prediction model with MSE, MAE, and \(R^2\) of 1.046E-5, 0.0026 and 0.65, respectively. Their experimental results demonstrated a 0.5\% to 2\% PUE reduction. Zhao et al. \cite{b13} proposed a GRU-based model for PUE prediction, leveraging historical data on energy consumption. They applied RFECV to select most important 34 features out of 144 energy-related features. They compared GRU with other models such as LSTM, ANN, SVM, and Linear SVM, and demonstrated that GRU based PUE prediction model outperformed all the other models on MSE, MAE, and \(R^2\) evaluation metrics.

\section{Proposed Approach}

In this section, the proposed framework for PUE prediction which involves data generation using EnergyPlus Platform, data preprocessing, selecting high priority features using RFECV, and developing a BiGRU-based deep learning model, with the aim to achieve high prediction accuracy of PUE is presented. The flowchart depicting the steps involved in the development of the PUE prediction model is presented in Fig.~\ref{fig1}.

\subsection{Data set generation}\label{AA}
\begin{figure}
    \centering
    \includegraphics[width=1\linewidth]{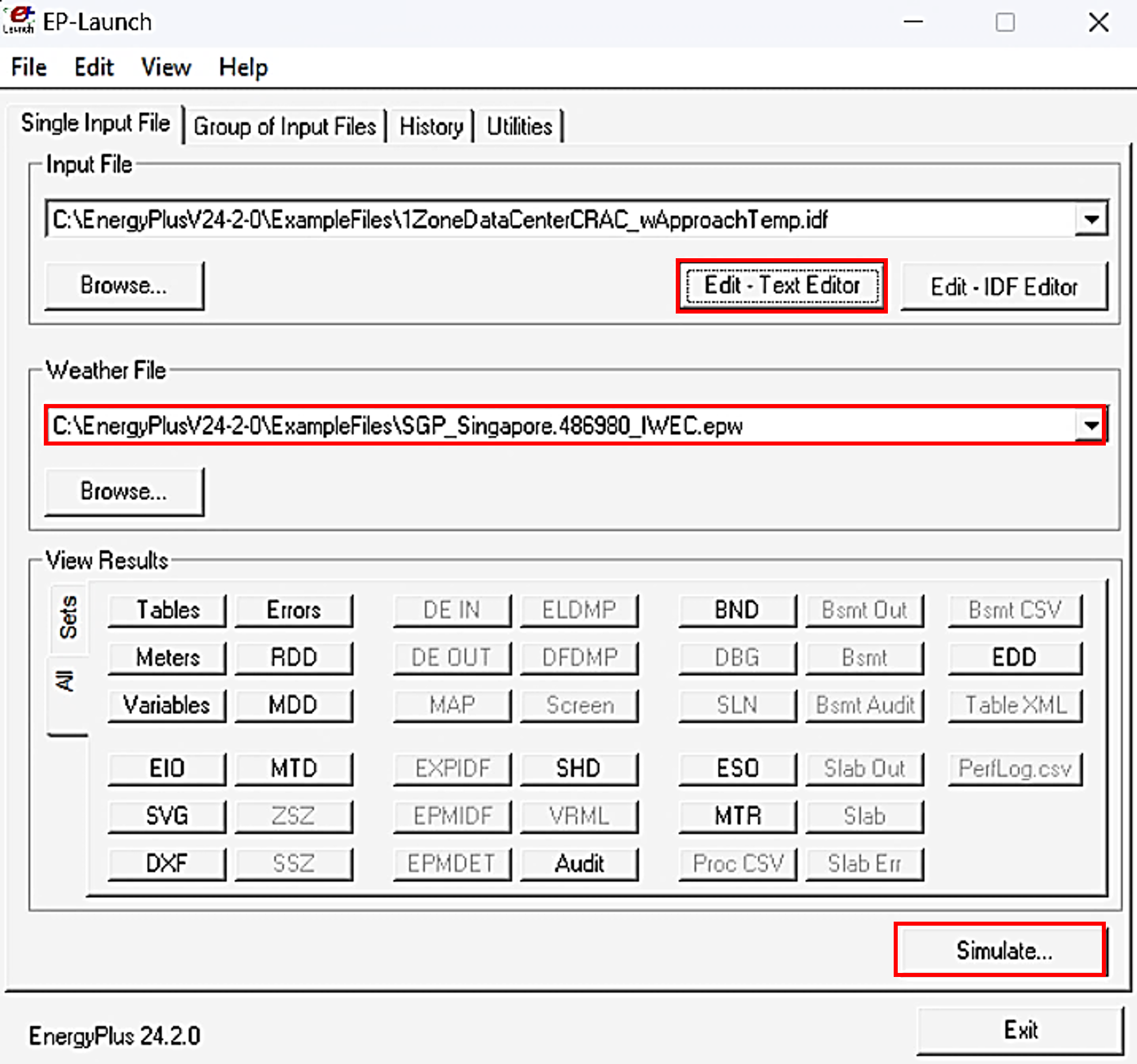}
    \caption{EP-Launch setup with Singapore IWEC weather file.}
    \label{fig2}
\end{figure}

The simulation software used for data generation is EnergyPlus v24.2.0 \cite{b18}. We launched EP-Launch (refer Fig.~\ref{fig2}) to use the EnergyPlus example file with hourly data which is the default input file. We added 2009 Singapore Weather conditions available with EnergyPlus Weather Data website as the input weather file. We made necessary changes in the IDF file using Edit - Text Editor. This includes changing the weather details based on the Singapore weather file, giving the time step as 6 and replacing all `hourly' with `Time Step' to get data every 10 minutes, and setting the start and end date of simulation as January 1 and December 31 to run the simulation for the duration of 1 year. By following this procedure, we collected year-long data having 52,560 data samples with 117 features (excluding timestamps) at 10-minute intervals.

\subsection{Normalization}
The 117 features are measured in different units such as Celsius, percentage, Watt, etc., resulting in widely varying ranges across features. Thus, the features with larger values can have a higher influence on the prediction model than the features with smaller values.

Normalization addresses this problem by bringing all features into a common range. This ensures that every variable contributes proportionately to the model training. Using min-max scaling, all features were normalized in the [0,1] range. We used the MinMaxScaler function from the Python library Scikit-learn with a feature range of [0,1].

\subsection{Feature Selection}
One of the main challenges of DCs is choosing the important features from hundreds of features. Using 117 features to train the PUE prediction model increases the problem complexity, and not all features are important. Therefore, we need to minimize the feature set to a minimum subset with the most important features. Many articles \cite{b9}, \cite{b10}, \cite{b12} directly indicate the features they used. They mention that the features are selected based on expert advice and do not use any feature selection method.
Contrary to the aforementioned works, in our work, a feature selection method named, Recursive Feature Elimination with Cross Validation (RFECV) \cite{b19} is utilized. RFECV removes the less important features one by one to achieve the best features that can be utilized to train the model. Cross validation ensures that these selected features apply to all parts of the data. The final output is the smallest set of features that gives the best performance.

The steps involved in RFECV (Algorithm~\ref{algo:1}) are as follows. Initially, the XGBoost model is trained with all features. Each feature is given a feature importance score, and the less important ones are removed in each iteration. Then, the model is retrained on the remaining features, and Cross Validation verifies the performance.

The parameters in RFECV algorithm includes:
\begin{itemize}
    \item Estimator: XGBoost is used to evaluate the feature importance. 
    \item Step: Number of features to remove at each iteration. We set the step size as 1.
    \item CV: Number of equal parts into which the data is divided, which in our case is 5. One part is for testing, and the remaining parts are for training. The testing part is different at each iteration.
    \item Verbose: Level of output log detail.
    \item Scoring: To evaluate the model’s performance. Since, PUE prediction is a regression problem, we chose MSE. 
\end{itemize}

The XGBoost estimator uses level-wise tree growth where all levels of the tree are split and evaluated, and residuals are calculated before passing to the next level. It does regularization to prevent overfitting and is more balanced, but slower. The hyperparameters of XGBoost includes learning rate, number of trees, and maximum tree depth. Varying this selects a different group of important features with slightly differing MSE.

\begin{algorithm} 
\caption{Feature Selection Using RFECV with XGBoost}
\begin{algorithmic}[1]
\STATE Define hyperparameters:
\begin{itemize}
    \item Learning rates, $LR = \{0.5, 0.75, 0.1, 0.075, 0.05\}$
    \item Number of estimators, \texttt{\textit{n\_estimators =} \{50, 100, 150, 200, 250\}}
    \item Maximum tree depth, $max\_depth = \{3, 6, 9, 12\}$
\end{itemize}
\FOR{each $lr \in LR$, $n \in n\_estimators $, $m \in max\_depth$}
    \STATE Train XGBoost with RFECV for feature selection.
    \STATE Evaluate model (MSE) on test data.
\ENDFOR
\STATE Save results: hyperparameters, MSE, selected features
\end{algorithmic}
\label{algo:1}
\end{algorithm}

  Though Zhao et al. \cite{b13} used RFECV for Feature Selection, the effect of different sets of selected features due to changes in the hyperparameter setting is not considered. Here, we select 6 groups of selected features with the least MSE, as lower loss from the XGBoost model does not necessarily mean that these features will perform the best with respect to the final PUE prediction model. This is because XGBoost does not capture the complex interdependencies, but is good enough to select important features. Thus, comparing the performance of these sets of selected features on the final BiGRU-based PUE prediction model is the only reliable method to determine the optimal features.
  
\begin{figure}
    \centering
    \includegraphics[width=1\linewidth]{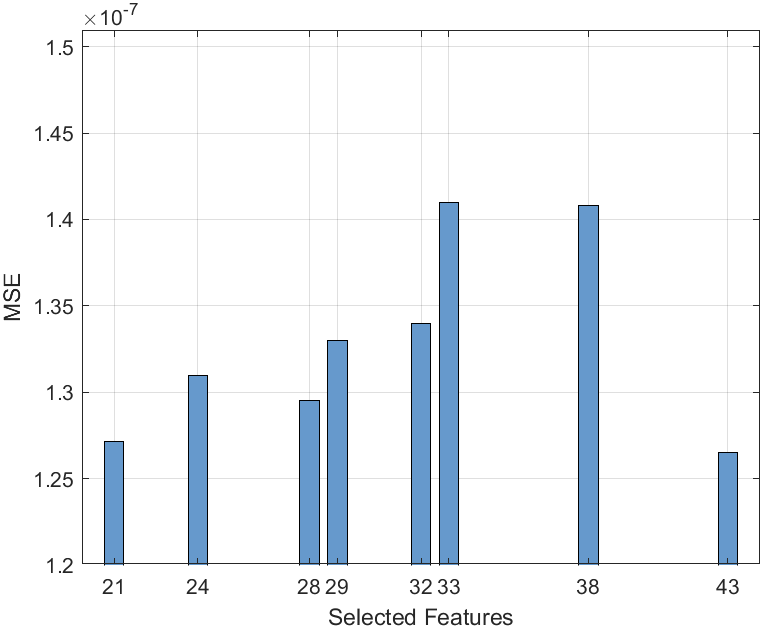}
    \caption{Number of selected features using RFECV vs. MSE}
    \label{fig3}
\end{figure}

  Table~\ref{tab:1} shows the number of selected features \{21, 28, 29, 32, 33, 43\} with the least MSE values for different hyperparameter settings of XGBoost in RFECV. Fig~\ref{fig3} visually illustrates the MSE values for different number of selected features {21, 28, 29, 32, 33, 43}. As an example, Table~\ref{tab:2} summarizes the 21 selected features.

\begin{table}
\caption{Selected Features for XGBoost in RFECV}
\begin{center}
\resizebox{\columnwidth}{!}{
\begin{tabular}{|c|c|c|c|c|}
\hline
\textbf{Learning} & \textbf{Number of} & \textbf{Maximum} & \textbf{MSE} & \textbf{Number of} \\ 
\textbf{Rate} & \textbf{Trees} & \textbf{Tree Depth} & & \textbf{Selected Features} \\
\hline
0.1  & 100  & 6  & $1.27169 \times 10^{-7}$  & 21 \\
0.075 & 150 & 6  & $1.30993 \times 10^{-7}$  & 24 \\
0.05 & 200  & 6  & $1.29504 \times 10^{-7}$  & 28 \\
0.1  & 100  & 12 & $1.33014 \times 10^{-7}$  & 29 \\
0.075 & 100 & 6  & $1.34 \times 10^{-7}$     & 32 \\
0.05 & 200  & 12 & $1.40961 \times 10^{-7}$  & 33 \\
0.05 & 150  & 6  & $1.40778 \times 10^{-7}$  & 38 \\
0.05 & 200  & 9  & $1.2649 \times 10^{-7}$   & 43 \\
\hline
\end{tabular}
}
\label{tab:1}
\end{center}
\end{table}

\begin{table*}[ht!]
\caption{Selected 21 features with description}
\begin{center}
    \renewcommand{\arraystretch}{1.2} 
    {\normalsize
    \begin{tabular}{|l|p{11cm}|c|}
        \hline
        \textbf{Category} & \textbf{Selected Features} & \textbf{Unit} \\
        \hline
        \multirow{3}{*}{\textbf{Environment}} 
            & Site Outdoor Air Drybulb Temperature & \(^{\circ}\text{C}\) \\
            & Site Outdoor Air Dewpoint Temperature & \(^{\circ}\text{C}\) \\
            & Site Outdoor Air Wetbulb Temperature & \(^{\circ}\text{C}\) \\
        \hline
        \multirow{4}{*}{\textbf{Data Center Servers}} 
            & ITE CPU Electric Power & W \\
            & ITE Fan Electric Power & W \\
            & ITE UPS Electric Power & W \\
            & ITE Standard Density Air Volume Flow Rate & m\(^3\)/s \\
        \hline
        \multirow{7}{*}{\textbf{Cooling System}} 
            & EMS:CRAC Total System Power & W \\
            & EMS:CRAC Net Sensible Capacity & W \\
            & EMS:CRAC SCOP & None \\
            & EC Plug Fan 1: Fan Electric Power & W \\
            & Main Cooling Coil 1: Cooling Coil Electric Power & W \\
            & Main Cooling Coil 1 Outlet Node: System Node Temperature & \(^{\circ}\text{C}\) \\
            & Main Cooling Coil 1 Outlet Node: System Node Setpoint Temperature & \(^{\circ}\text{C}\) \\
        \hline
        \multirow{7}{*}{\textbf{Basic Facility}} 
            & Main Zone: Zone Air Heat Balance Surface Convection Rate & W \\
            & Main Zone: Zone Air Heat Balance System Air Transfer Rate & W \\
            & Main Zone: Zone Air Heat Balance Air Energy Storage Rate & W \\
            & Whole Building: Facility Total HVAC Electricity Demand & W \\
            & Supply Inlet Node: System Node Dewpoint Temperature & \(^{\circ}\text{C}\) \\
            & Supply Outlet Node: System Node Temperature & \(^{\circ}\text{C}\) \\
            & Main Zone Inlet Node: System Node Temperature & \(^{\circ}\text{C}\) \\
        \hline
    \end{tabular}}
    \label{tab:2}
    \end{center}
\end{table*}

\subsection{PUE Prediction model}

RNNs were first used in language models to capture long-term dependencies. RNN can handle variable-length sequential data by having a recurrent hidden state which remembers past data \cite{b20}. Since the parameters contributing to PUE are time-dependent and sequential, RNN is the most suitable for developing PUE prediction model. However, RNN has the issue of gradients vanishing. To solve this, LSTM and GRU were introduced \cite{b13} with forget gates, which help memory cells to decide what information to forget. BiGRU is an improved version of GRU that uses both past and future data. In the upcoming part, we share further details on GRU and BiGRU models.

\subsubsection{\textbf{GRU}}
GRU is similar to LSTM but contains only 2 gates: a reset gate and an update gate, combining the functions of forget and input gates of LSTM. Their roles are as follows:
 
\begin{itemize}
    \item Reset gate (\(r_t\)): It tells the extent to which hidden information at the previous time is forgot, with a value of [0,1]. If it is closer to 1, then, more of past data is retained.

    \item Update gate (\(z_t\)): It determines what information is to be discarded and what is to be added, with the value of [0,1]. If closer to 1, then more data of current state is from past memory than the current input \cite{b21}.
\end{itemize}

\begin{equation}
r_t = \sigma\left( \mathbf{W_r} \cdot h_{t-1} + \mathbf{U_r} \cdot x_t + \mathbf{b_r} \right)
\label{eq1}
\end{equation}
\begin{equation}
z_t = \sigma\left( \mathbf{W_z} \cdot h_{t-1} + \mathbf{U_z} \cdot x_t + \mathbf{b_z} \right)
\label{eq2}
\end{equation}
\begin{equation}
\hat{h}_t = \tanh\left( \mathbf{W_h} \cdot (r_t \odot h_{t-1}) + \mathbf{U_h} \cdot x_t + \mathbf{b_h} \right)
\label{eq3}
\end{equation}
\begin{equation}
h_t = (1 - z_t) \cdot h_{t-1} + z_t \cdot \hat{h}_t
\label{eq4}
\end{equation}

The above represents the computation of each gate \cite{b22}. \( h_t \) and \(\hat{h}_t\) are the candidate and output hidden states of the GRU at time \( t \). \( \mathbf{W_r} \), \( \mathbf{W_z} \), and \( \mathbf{W_h} \) are the weight matrices for the reset gate, update gate, and candidate hidden state, respectively. The matrices \( \mathbf{U_r} \), \( \mathbf{U_z} \), and \( \mathbf{U_h} \) are the weight matrices for input \( x_t \) \cite{b23}. \( \mathbf{b_r} \), \( \mathbf{b_z} \), and \( \mathbf{b_h} \) are the biases.

\subsubsection{\textbf{BiGRU}}

 In BiGRU, two GRUs are stacked to model a sequence in both directions. Both the forward GRU (\(h_t(f)\)) and the backward GRU (\(h_t(b)\)) generate two hidden states at each time step. These hidden states are added from both sides to produce the complete representation of the input data at time \textit{t} \cite{b21}.

\begin{equation}
h_t = h_t(f) + h_t(b)
\label{eq5}
\end{equation}

Since BiGRU accounts for data from both directions, it may perform better than GRU.

\section{Results and Discussion}
The experiments were conducted using Python. We imported Pandas for reading and writing data from CSV file, NumPy for numerical calculations. For feature selection using RFECV, data splitting, normalization, and performance evaluation using MSE, MAE, and \(R^2\) metrics, we employed Scikit-learn. XGBoost was included for deploying it as the estimator in RFECV. To implement the GRU and BiGRU-based deep learning networks, PyTorch's torch.nn module was utilized. To optimize and train these models, the torch.optim tool was used. In order to enhance training efficiency, we executed the code on a system configured with CUDA version 12.7 for GPU acceleration. MATLAB was used for data visualization.

\begin{figure*}
    \centering
    \includegraphics[width=1\linewidth]{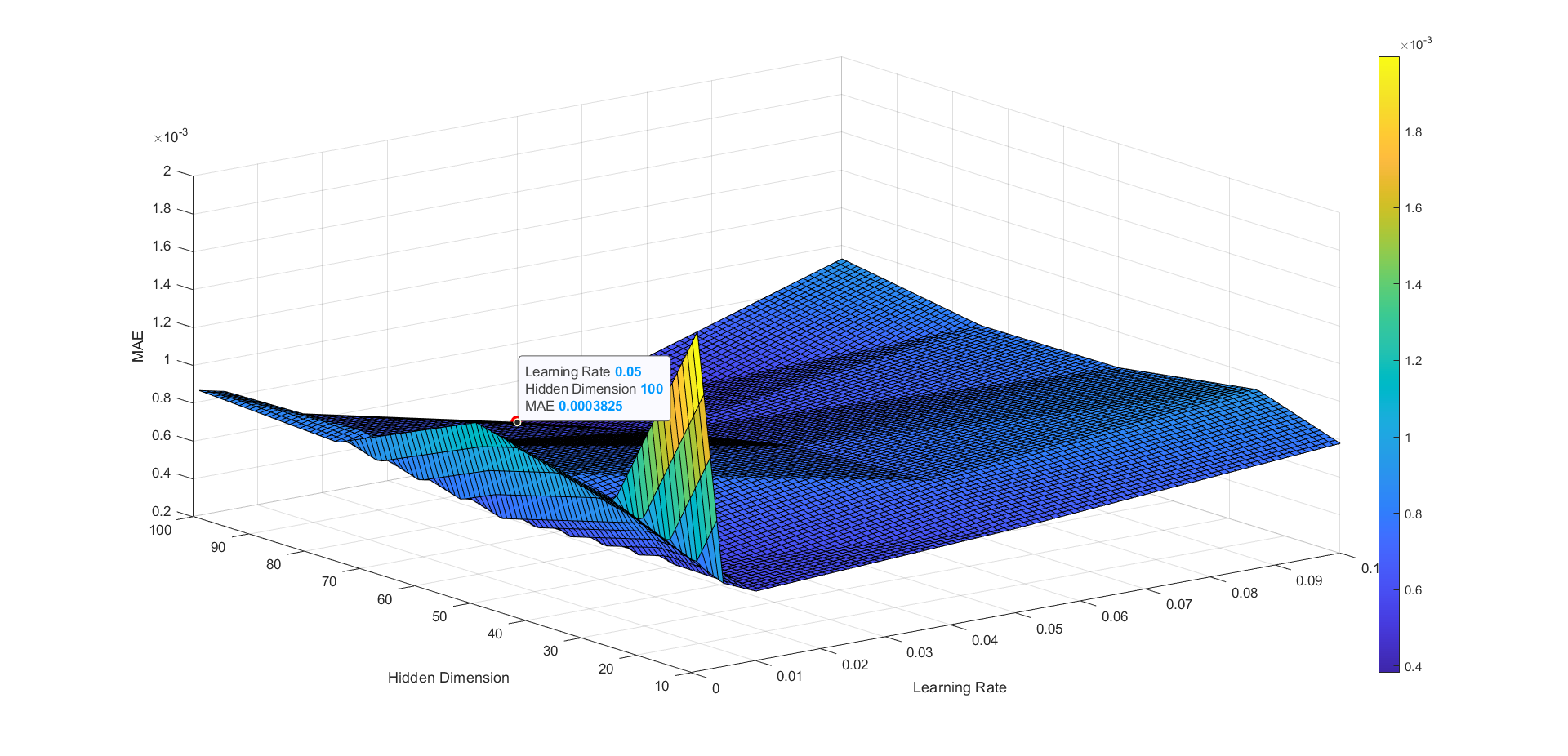}
    \caption{BiGRU Hyperparameter tuning for 21 selected features}
    \label{fig4}
\end{figure*}

We developed the BiGRU-based PUE prediction model. Using RFECV, we obtained \{21, 28, 29, 32, 33, 43\} set of selected features as discussed in Section III.C. Thereafter, we conducted the hyperparameter tuning (Algorithm~\ref{algo:2}) individually to obtain the optimal configuration with MSE as the loss function. Thus, we obtained an improved BiGRU-based PUE prediction model with optimal hyperparameter settings. Fig.~\ref{fig4} depicts the relationship between the hyperparameters and MAE during hyperparameter tuning of BiGRU with 21 features.

\begin{algorithm}
\caption{Optimized BiGRU with hyperparameter tuning}
\begin{algorithmic}[1]
\STATE Define hyperparameters:
\begin{itemize}
    \item Number of layers, $L = \{1, 2, 3\}$
    \item Hidden dimensions, $H = \{10, 25, 50, 75, 100\}$
    \item Learning rates, $LR = \{0.001, 0.005, 0.01, 0.05, 0.1\}$
\end{itemize}
\STATE Initialize:
\begin{itemize}
    \item best loss = $\infty$
    \item best params = $\emptyset$
\end{itemize}
\FOR{each $l \in L$, $h \in H$, $lr \in LR$}
    \STATE Initialize BiGRU model, loss function (MSE), optimizer (Adam)
    \FOR{each epoch = 1 : 4000}
        \STATE Perform forward pass, compute loss
        \IF{epoch is a multiple of 500}
            \STATE Evaluate performance (MSE, MAE, R²)
        \ENDIF
        \IF{loss $<$ best loss}
            \STATE Update best params, save model
        \ENDIF
    \ENDFOR
\ENDFOR
\STATE Save best model and metrics
\STATE Output: Best configuration, best MSE, MAE, R²
\end{algorithmic}
\label{algo:2}
\end{algorithm}

\begin{table*}[ht]
\centering
\caption{BiGRU Performance with Optimal Hyperparameter Settings for Selected Features}
\begin{tabular}{|c|c|c|c|c|c|c|c|}
\hline
\textbf{Selected} & \textbf{Number of} & \textbf{Hidden} & \textbf{Learning} & \textbf{Epochs} & \textbf{MSE} & \textbf{MAE} & \textbf{$R^2$} \\  
\textbf{Features} & \textbf{Hidden Layers} & \textbf{Dimension} & \textbf{Rate} & & & &\\ \hline  
\rowcolor{gray!30}
21 & 1 & 100 & 0.05 & 3500 & $5.6342 \times 10^{-7}$ & 0.0003825 & 0.99673 \\ 
24 & 3 & 10 & 0.005 & 4000 & $5.8791 \times 10^{-7}$ & 0.0004863 & 0.996588 \\ 
28 & 3 & 10 & 0.01 & 4000 & $6.9211 \times 10^{-7}$ & 0.0004899 & 0.995983 \\ 
29 & 3 & 25 & 0.005 & 4000 & $1.0202 \times 10^{-6}$ & 0.0005987 & 0.994078 \\ 
32 & 2 & 10 & 0.01 & 4000 & $9.1791 \times 10^{-7}$ & 0.0005336 & 0.994672 \\ 
33 & 3 & 25 & 0.01 & 3000 & $8.4943 \times 10^{-7}$ & 0.0005467 & 0.99507 \\ 
38 & 3 & 50 & 0.005 & 4000 & $9.9167 \times 10^{-7}$ & 0.0005689 & 0.994244 \\
43 & 3 & 50 & 0.005 & 3000 & $9.871 \times 10^{-7}$ & 0.0005683 & 0.99427 \\ \hline
\end{tabular}
\label{tab:3}
\end{table*}

\begin{table*}[ht]
\centering
\caption{GRU Performance with Optimal Hyperparameter Settings for Selected Features}
\begin{tabular}{|c|c|c|c|c|c|c|c|}
\hline
\textbf{Selected} & \textbf{Number of} & \textbf{Hidden} & \textbf{Learning} & \textbf{Epochs} & \textbf{MSE} & \textbf{MAE} & \textbf{$R^2$} \\  
\textbf{Features} & \textbf{Hidden Layers} & \textbf{Dimension} & \textbf{Rate} & & & &\\ \hline  
21 & 2 & 50 & 0.01 & 3500 & $6.1698 \times 10^{-7}$ & 0.0004928 & 0.996419 \\ 
24 & 3 & 75 & 0.01 & 3000 & $6.8952 \times 10^{-7}$ & 0.0005005 & 0.995998 \\ 
\rowcolor{gray!30}
28 & 3 & 25 & 0.01 & 3500 & $6.2282 \times 10^{-7}$ & 0.0004181 & 0.996385 \\
29 & 3 & 25 & 0.01 & 4000 & $6.8262 \times 10^{-7}$ & 0.0004487 & 0.996038 \\ 
32 & 2 & 50 & 0.005 & 3500 & $6.7597 \times 10^{-7}$ & 0.0004768 & 0.996076 \\ 
33 & 2 & 50 & 0.01 & 3000 & $8.3053 \times 10^{-7}$ & 0.0005404 & 0.995179 \\ 
38 & 3 & 25 & 0.01 & 3500 & $7.3268 \times 10^{-7}$ & 0.0005162 & 0.995747 \\ 
43 & 1 & 25 & 0.01 & 4000 & $7.802 \times 10^{-7}$ & 0.0006026 & 0.995471 \\ \hline
\end{tabular}
\label{tab:4}
\end{table*}

\begin{figure*}
    \centering
    \includegraphics[width=1\linewidth]{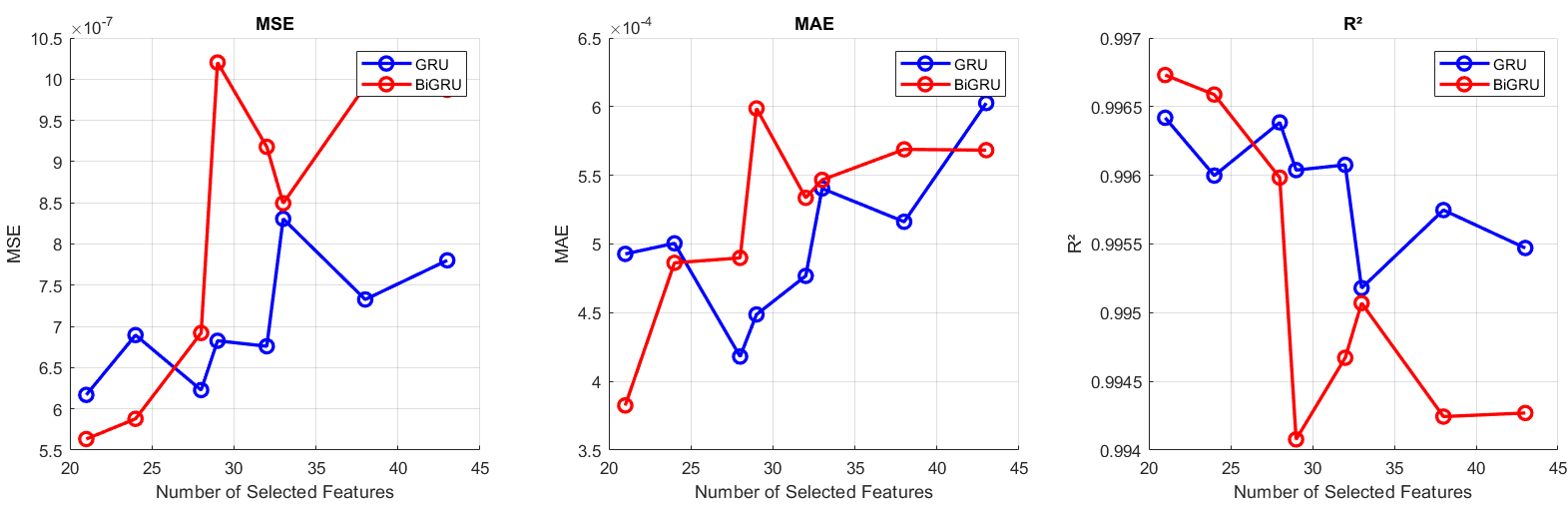}
    \caption{BiGRU vs GRU performance over different set of selected features}
    \label{fig:grubigru}
\end{figure*}

The optimal hyperparameter settings and the performance metrics of each set of features are provided for both BiGRU (Table~\ref{tab:3}) and GRU (Table~\ref{tab:4}) based PUE prediction models. Fig.~\ref{fig:grubigru} shows the performance comparison for BiGRU and GRU over different sets of selected features with respect to the three metrics. We can observe that the proposed BiGRU model achieved more accurate predictions than GRU, with the lowest MSE and MAE values and the highest \(R^2\) score, which indicates minimal error and superior predictive performance using only 21 features.

The following observations are made from the final results:
\begin{itemize}
    \item Table~\ref{tab:3} shows that the best performance for BiGRU-based PUE prediction model is obtained with 21 features for the model architecture of 1 hidden layer and 100 hidden neurons at a learning rate of 0.05.
    \item Table~\ref{tab:4} indicates that the best performance for GRU-based PUE prediction model is obtained with 28 features for the model architecture of 3 hidden layers and 25 hidden neurons at a learning rate of 0.01.
    \item We can observe that different set of features have different optimal hyperparameter configurations.
    \item From Fig~\ref{fig:grubigru}, we observe that the overall best performance is obtained with BiGRU at 21 features as it has the least MAE and MSE values, and the highest \(R^2\) value.
    \item Fig~\ref{fig:grubigru} highlights the significance of considering different selected features in the design of the Bi-GRU as well as GRU models, illustrating that it is not guaranteed exactly which and how many number of features may result in the best performance of the models.
\end{itemize}

\section{Conclusion}
In this paper, we developed a prediction model for PUE, based on BiGRU, and extensively compared it with the GRU model. We used 52,560 samples of dataset generated by EnergyPlus simulation of a data center in Singapore and the RFECV approach with XGBoost to identify the most relevant features for prediction. Different parameter settings of the XGBoost model resulted in different number of selected features, and we chose only those with the lowest MSE. We then developed the BiGRU model for predicting PUE with these chosen feature sets and found the optimal hyperparameters for each by hyperparameter tuning. As a result, we obtained an optimized BiGRU-based PUE prediction model. We followed the same for the GRU model. We observed that the BiGRU model has higher accuracy in predicting PUE using 21 features with the lowest MAE of 0.00038, while GRU performed the best with 28 features with an MAE of 0.0004181. Thus, the BiGRU-based PUE prediction model performs better than the GRU-based model by lowering the input dimensionality with less number of input selected features while achieving improved PUE prediction accuracy. This demonstrates the effectiveness of the BiGRU design and RFECV feature selection with hyperparameter tuning approaches in predicting PUE of data centers. This method could be extended to real-time PUE prediction scenarios in future research to help identify the best feature settings that will contribute to energy efficient data centers.

\vspace{12pt}
\end{document}